\documentclass[5p]{elsarticle}

\usepackage{hyperref}

\usepackage{graphicx}
\usepackage{amsmath,amssymb}
\usepackage{multirow}
\usepackage{algorithm}
\usepackage[]{algorithmic}
\usepackage{siunitx}

\usepackage[T1]{fontenc}
\usepackage[utf8]{inputenc}
\usepackage{soulutf8}

\usepackage{color}
\usepackage{soul}

\usepackage[capitalise, noabbrev]{cleveref}
\usepackage{subcaption}
\usepackage{caption}

\journal{Information Fusion}

\begin{document}
	
    \begin{frontmatter}
		
		\title{InMyFace: Inertial and Mechanomyography-Based Sensor Fusion for Wearable Facial Activity Recognition}
  
        \author[DFKI]{Hymalai Bello \corref{mycorrespondingauthor}}
		\cortext[mycorrespondingauthor]{Corresponding author}
		\ead{hymalai.bello@dfki.de}
		\author[DFKI]{Luis Alfredo Sanchez Marin}
        \author[DFKI,TUKaiserslautern]{Sungho Suh}
        \author[DFKI,TUKaiserslautern]{Bo Zhou}
		\author[DFKI,TUKaiserslautern]{Paul Lukowicz}
		
		\address[DFKI]{German Research Center for Artificial Intelligence (DFKI), 67663 Kaiserslautern, Germany}
		\address[TUKaiserslautern]{Department of Computer Science, RPTU Kaiserslautern-Landau, 67663 Kaiserslautern, Germany}

	\begin{abstract}
        Recognizing facial activity is a well-understood (but non-trivial) computer vision problem. 
        However, reliable solutions require a camera with a good view of the face, which is often unavailable in wearable settings. 
        Furthermore, in wearable applications, where systems accompany users throughout their daily activities, a permanently running camera can be problematic for privacy (and legal) reasons. 
        This work presents an alternative solution based on the fusion of wearable inertial sensors, planar pressure sensors, and acoustic mechanomyography (muscle sounds). 
        The sensors were placed unobtrusively in a sports cap to monitor facial muscle activities related to facial expressions.  
        We present our integrated wearable sensor system, describe data fusion and analysis methods, and evaluate the system in an experiment with thirteen subjects from different cultural backgrounds (eight countries) and both sexes (six women and seven men).
        In a one-model-per-user scheme and using a late fusion approach, the system yielded an average F1 score of 85.00\% for the case where all sensing modalities are combined. 
        With a cross-user validation and a one-model-for-all-user scheme, an F1 score of 79.00\% was obtained for thirteen participants (six females and seven males). 
        Moreover, in a hybrid fusion (cross-user) approach and six classes, an average F1 score of 82.00\% was obtained for eight users.
        The results are competitive with state-of-the-art non-camera-based solutions for a cross-user study. 
        In addition, our unique set of participants demonstrates the inclusiveness and generalizability of the approach.
        \end{abstract}
		
	\end{frontmatter}
	

\section{Introduction}
\label{introduction}
\begin{figure*}[htbp]
    \centering
    \includegraphics[width=0.8\textwidth]{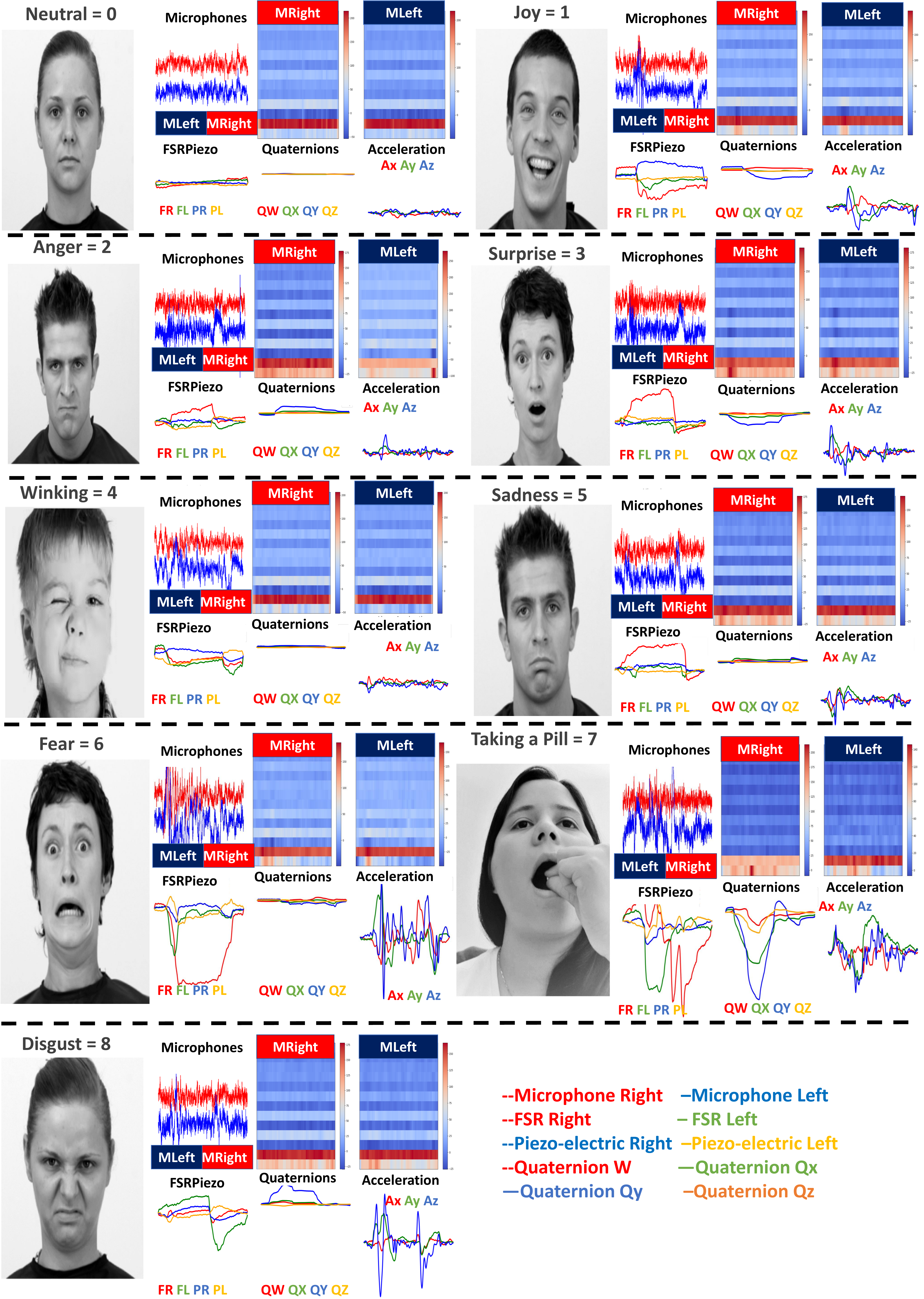}
    \caption{Facial Muscle Activities Dictionary with Sensor Signal Examples; 7 Facial Expressions from Warsaw Photoset \cite{Warsaw} and 2 Gestures from \cite{bello2020facial}. Two Channel Raw Audio Data, Thirteen Mel Frequency Cepstral Coefficients (Two Channel Audio-Monophonic), Force Sensitive Resistor, Piezoelectric Film, Orientation and Acceleration.}
    \label{fig:FacialSignals}
\end{figure*}

Facial expressions are a key means of communication in human interactions; they are sometimes more informative than explicit verbal communication \cite{FacialNonverbal,NonVerbaleffects}. 
They affect our social intelligence and ability to create interpersonal emotional connections \cite{FacialSocialIntelligence}.  
Consequently, monitoring facial expressions has been investigated in many pervasive computing applications. 
Examples include novel human-computer interfaces \cite{RobotHumanFacial, HCIINDModel}, learning feedback \cite{FacialCamKidClassroom}, and recognizing a car's driver's fatigue or mood \cite{tiredness1,tiredness2}.

Facial expression recognition is a complex problem for several reasons. 
First and foremost, its large interpersonal variability \cite{FacialExpInfo}.
It is influenced by cultural background \cite{FacialJapanCult,FacialCult2}, age, sexes \cite{FacialAndGender}, race, and other person-specific characteristics, leading to a user-dependent solution in many related works \cite{HCIINDModel} (see Table \ref{table:comparison}).  
Recently, in \cite{facialGenderBias}, the authors have formally studied the impact of a sex-balanced dataset on the fairness of the results for facial expression recognition (happiness, sadness, surprise, fear, disgust, and anger). 
They conclude that training with the mixed dataset achieves the best results in all cases. 
Furthermore, fairness is compromised in training with a highly biased dataset, especially when classifying particular expressions.
In addition, different expression categories may have only minor differences (e.g., anger and sadness, as depicted in \cref{fig:FacialSignals}).

Cameras are the most widely used solution to recognize expressions with an expected accuracy of more than 90.00\% \cite{CameraEmotions, CameraHeadMounted, CfaceCam}.
However, in wearable applications, cameras are not always a suitable solution. 
First, placing a camera on the user's body in an unobtrusive, suitable for everyday use, and simultaneously providing a good view of the entire face is difficult at best. 
Light conditions and occlusion can also be a problem.  
In addition, as the example of GoogleGlass has shown, having a permanently body-worn camera in everyday situations can be socially awkward and even illegal in some countries.  
Computer vision-based methods also tend to have a larger memory footprint and power consumption than non-visual sensor-based solutions, such as the one presented in \cite{PiezoFacialNature}.

There has been significant interest in the wearable community in non-visual sensing approaches for face monitoring. 
However, as related work describes, most systems struggle to balance wearability constraints with the need to acquire clean and informative signals. 
The core of the problem is that facial muscle movements involve large areas of the face that sensors cannot cover unobtrusively. 
Additionally, sensor placement is limited to areas such as: around glasses, on the head (e.g., under a cap), and around the ears.  
The challenge is to use those limited areas to capture information related to the overall facial activity. 
Approaches that were previously studied include inertial measurement unit (IMU) \cite{EarbudsImu}, ultrasound \cite{EarCanalUltra}, and photo reflective sensors \cite{PhotoReflectiveGlass}. 
Acoustic sensing for face monitoring can be more compact, with a smaller data footprint and computational requirements, and lower power consumption compared to cameras \cite{EarCanalUltra, EarIO}.
Nevertheless, the privacy concern of camera/audio-based design still limits user acceptance.

A standard method for quantifying muscle movements is their electrical activation pattern, sensed by electromyography (EMG).
EMG measurements rely on detecting weak electrical signals from the body. 
This technique requires a stable electrical connection to the skin surface, typically implemented using micro-needles embedded in the muscles or at least wet electrodes. 
Both options can be a challenge for use in everyday wearable settings \cite{EMGMMG}. 
On the other hand, mechanomyography (MMG) measures the mechanical signals resulting from the lateral movement of muscle fibers at low-frequency \cite{MMGDefinition,MMGDefinition2}.
MMG does not need an electrical connection with the skin and can be detected by various types of sensors, such as IMU, piezoelectric, microphones, pressure sensors, laser sensors, and many others \cite{MMGSensorChoice}. 
In \cite{bello2020facial}, facial muscle acoustic mechanomyography (AMMG) has been proposed.
The authors used stethoscope microphones distributed around the temporalis, frontalis, and masseter muscles to detect ten facial expressions.
An F1 score of 54.00\% for the cross-user case (eight volunteers) shows promising results for a passive, privacy-aware method to detect facial muscle movements. 
However, it is still not a pleasant solution to wear and is far from being a portable option. 
In \cite{Expressure}, a pressure mechanomyography (PMMG) solution has been presented with 38.00\% accuracy in classifying seven Warsaw Photoset facial expressions \cite{Warsaw}. 
The authors showed the potential of an unobtrusive, passive, textile, headband-based design for monitoring facial muscle activity.

Alternatively, multimodal approaches have been studied to exploit the limitation of independent sensing modalities. 
The idea is to combine multiple data sources with complementary information to reduce ambiguity, add completeness to the situation being studied ("gain in representation"), improve the signal-to-noise ratio(assuming independent error sources), and increase the confidence in the model decision("gain in robustness"). 
In \cite{FaceCommandGlasses}, the authors proposed a combination of IMU and 16 optical sensors in smart glasses to detect eight temporal facial gestures. 
Obtaining F1 score results of 91.10 \% for the case of one model per person for the recognition of facial action units (AUs: AU12, AU27, LP, AU1+2, AU4, AU43, AU46R, AU61) \cite{ekmanAU}. 
Optical sensors for facial muscle movements limit the system to stable light conditions. 
In \cite{KissglassIMUOculo}, another solution based on the fusion of IMU and electrooculography is presented to recognize kissing gestures, obtaining an accuracy of 74.33\% in a cross-user scheme. 
IMU is the common basis in the multimodal scheme, so it is a source of information to be considered.

This work proposes a multimodal system that fuses complementary sensing modalities, such as inertial sensing IMU (orientation and acceleration), AMMG, and  PMMG (force-sensitive resistor and piezoelectric film),  integrated into a sports cap accessory. 
The facial expressions to be evaluated come from the Warsaw Photoset (seven expressions) and \cite{bello2020facial} (two facial movements); thus, it will be simpler to compare with future solutions, see \cref{fig:FacialSignals}. 

In summary, \textbf{our contributions} are: 
\begin{itemize}
    \item We present a multimodal sensing alternative for facial muscle motion monitoring based on inertial, planar pressure and acoustic sensors distributed in a minimally obstructive wearable accessory (sports cap). Furthermore, the idea can be adapted and potentially unobtrusively integrated into other head-worn platforms (e.g., glasses and headbands) without compromising the sensing capability.
    \item We adopt a modular multimodal fusion method based on sensor-dependent neural networks using a late fusion approach with a low memory footprint ($\leq$ 2 MB) to simplify the future deployment of the idea in wearable/embedded devices with tiny dimensions and reduce memory (4 MB to 16 MB Flash). 
    \item We conduct a user study with thirteen participants from diverse cultural backgrounds (eight countries) and both sexes (six women and seven men). 
    Our unique dataset demonstrates the inclusiveness and generalizability of the results.
    For the evaluation, we use the Warsaw Photoset \cite{Warsaw} plus two facial gestures from \cite{bello2020facial}.
    \item We evaluate the system using a hybrid fusion approach with locally connected inception blocks with dimension reduction per sensing modality for the best eight imitators of the facial expression dictionary in \cref{fig:FacialSignals}, and six classes. 
\end{itemize}

The article is organized as follows.
\cref{sec:relatedwork} reviews related work on facial expression monitoring with wearables and mechanomyography (MMG). 
\cref{sec:approach} details the material and method, including the apparatus, experimental design, and information fusion approach. 
We show the experimental results and discuss our limitations in \cref{sec:results}. 
In \cref{sec:conclusion}, we conclude the article and make some remarks for the future.
	
\section{Related Work}
\label{sec:relatedwork}
\begin{table*}[!t]
\caption{Comparison with State-of-the-art Wearable Sensing Methods for Facial Activity Recognition}
\resizebox{\textwidth}{!}{
\setlength{\tabcolsep}{5pt}
\begin{tabular}{m{10em} m{10em} m{11em} m{10em} m{14em}}
\hline
Studies & Device& Expressions & Method & Performance \\
\hline
Ear-Canal(Active)\cite{EarCanalUltra}
& Ultrasonic in Earbud
& \textbf{21} Customized 
& SVM \textbf{user-dependent,  Users = 11}
& \textbf{62.5 \%} for 21 Facial Expressions and \textbf{90.00 \%} for 6
\\
ExpressEar(Passive)\cite{ExpressEar}
& IMU in Earbuds 
& \textbf{30} Action Units 
& One TCN \textbf{per user, Users = 12} 
& Accuracy \textbf{89.90 \%}
\\
Face Commands(Active)\cite{FaceCommandGlasses}
& 16 Optical sensors and IMU-Smart Glasses
& \textbf{8} Temporal gestures
& One CNN \textbf{per user, Users = 13}
& F1 \textbf{ 91.10 \%}
\\
Earbuds(Passive)\cite{EarbudsImu} 
& IMU on Ear
& \textbf{3} Head movements 
& Hierarchical Classification \textbf{user-dependent, Users = 21}
& F1 score for smile,talk an yawn \textbf{84.79 \%}
\\
FaceListener(Active)\cite{FaceListener}
& Ultrasound Headphones
&\textbf{7} Facial Expressions   
& One LSTM \textbf{per user, Users = 5}
& Accuracy \textbf{80.00 \%}
\\
Expressure(Passive)\cite{Expressure} 
& Textile Pressure in Headband
& \textbf{7} Facial Expressions
& SVM \textbf{cross-user,  Users = 20}
& Accuracy \textbf{38.10 \%}
\\
Microphones(Passive)\cite{bello2020facial} 
& 6 Stethoscope Microphones on Face
& \textbf{10} Facial Activities
& SVM  \textbf{cross-user,  Users = 8}
& F1 score \textbf{54.00 \%}
\\
FaceGlasses(Passive)\cite{Facerecglasses} 
& 2 Cameras in eyeglass shape device
& \textbf{7} Pseudo Face Image
& Comparison with stored images \textbf{user-dependent, Users = 6}
& Accuracy \textbf{87.50 \% neutral, 66.70 \%  happy and 71.40 \% surprise } 
\\
Neckface(Active)\cite{Neckface} 
& Infrared Cameras in necklace and neckband
& \textbf{8} Customized Facial Movements with Head Rotation
& CNN Real Time Tracking\textbf{ user-dependent, Users = 13}
& MAE \textbf{30.29} necklace and \textbf{25.61} neckband  
\\

C-Face(Passive)\cite{CfaceCam}
& 2 Ear mounted cameras
& \textbf{8} Facial Expressions (emojis)
& One BLSTM \textbf{per user, Users = 9} 
& Accuracy \textbf{88.60 \%}
\\

EmotGlass(Passive)\cite{EmotionCamEDAPPG}
& Camera, EDA and PPG 
& Quadrant of the Arousal–Valence Plane 
& SVM \textbf{user-dependent, Users = 20}
& Accuracy \textbf{76.09 \%}
\\
PhotGlasses(Active)\cite{PhotoReflectiveGlass}
& 17 Photo Reflective Sensors
& \textbf{8} Facial Expressions 
& One SVM \textbf{per user, Users = 8}
& Accuracy \textbf{78.10 \%}
\\
CanalSense(Passive)\cite{CanalsenseBaro}
& Barometer in earphones
& \textbf{11 Face Related Movements} 
& One Random Forest \textbf{per user, Users = 12}
& Accuracy \textbf{87.60 \%}
\\
KissGlass(Passive)\cite{KissglassIMUOculo}
& IMU and Electrooculography in Glasses
& \textbf{9} Kissing Gestures and Walking
& \textbf{Cross-user, Users = 5}
& Accuracy \textbf{74.33 \%}
\\
InfraredEar(Active)\cite{InfraredEar}
& IR sensor array on Ear Accessories
& \textbf{9} Facial Gestures
& One SVM \textbf{per user, Users = 5}
& F1-score \textbf{97.00 \%}
\\
CapGlasses(Active)\cite{Capglasses}
& Capacitive Channels in Glasses
& \textbf{12} Facial and Head Gestures
& One Random Forest \textbf{per user, Users = 10}
& Accuracy \textbf{89.60 \%}
\\
EarIO(Active)\cite{EarIO}
& Acoustic Sensing in Earphone 
& \textbf{9} Customized Facial Movements
& NN \textbf{user-dependent regression, Users = 12}
& MAE \textbf{25.90}
\\
\textbf{Our Approach Sportscap (Passive)}
& IMU, FSR, Piezo Electric and Microphones
& \textbf{See \cref{fig:FacialSignals}} 
& CNN \textbf{per user and cross-user, Users = 13}
& Per user F1-score \textbf{85.00 \%}, cross-user F1-score \textbf{79.00 \%} for 13 users and F1-score \textbf{82.00\%} for 8 users and six classes
\\

\hline
\label{table:comparison}
\end{tabular}}
\vspace{-20pt}
\end{table*}

This section examines previous research projects related to InMyFace in the scope of facial monitoring with wearables and pressure/audio-based mechanomyography. 
\vspace{-10pt}
\subsection{Facial Monitoring with Wearables}
The wearable community has explored facial expression monitoring with different sensing modalities embedded in accessories. 
The various sensing methods include cameras \cite{CameraEmotions, Neckface}, IMU \cite{KissglassIMUOculo, EarbudsImu}, light \cite{PhotoReflectiveGlass}, capacitive \cite{Capglasses}, piezo-electric \cite{PiezoGlasses}, electromyography (EMG) \cite{EMGperusquia}, mechanomyography \cite{Expressure, bello2020facial}, audio based solutions \cite{bello2020facial, FaceListener, MicArrayActiveFacialGlasses}, and many others. 
\cref{table:comparison} shows a comparison of the state-of-the-art approaches.

A popular accessory to deploy sensors is glasses. 
Recently, capacitive sensors have been embedded in glasses in \cite{Capglasses} to recognize 12 facial gestures and head movements.
They explored two ways to deploy the electrodes; one involves injecting a 12V AC signal into the body to elevate the body's potential and increase the signal-to-noise ratio, and the second one by generating a tremendous electric field close to the person's face/eyes to remove the ground dependency in capacitive sensing \cite{bello2021mocapaci, bellomove} and get good signals. 
Another active sensing approach in glasses is presented in \cite{PhotoReflectiveGlass}, using 17 photo-reflective sensors to classify eight facial expressions with results between 78.00-92.00\% per user.
Compact wearable cameras were deployed in front of the face using glasses in \cite{CfaceCam}. 
They focused on tracking facial muscle movement and generated a virtual avatar. 
Camera and photo-sensor-based methods share the sensitivity to the light condition.

In-ear wearable devices are a trend, with studies on facial expression recognition and the embedding of IMU \cite{EarbudsImu}, infrared \cite{InfraredEar}, electric field \cite{Earfieldsensing} and ultrasonic sensors in the ear canal \cite{EarCanalUltra}. 
A headphone-ultrasound device is proposed in \cite{FaceListener} and can track face deformation using the disturbances on the active-continuous sound signals being broadcast toward the face. 
A similar method to \cite{FaceListener} is proposed in \cite{EarIO}. 
The authors used their own designed earphones with pairs of speakers/microphones on both sides to propagate sound waves through the face. 
Based on the distortion on the microphones' inputs, they then tracked the skin deformations of customized facial movements. 
The results in \cite{EarIO} shows a promising future for a low-power and active audio system for facial movements tracking.

Active solutions could have negative consequences or long-term health effects, which physicians have not yet directly investigated. 
Some studies suggest the need for extensive research on standard technology used in wearables, such as radiation sources/waves around the body (e.g., WiFi and ultrasound) \cite{HealthUltraInfra, AudioVestibular, UltraExposure, RadiationFertility,MagExpMiscarriage}. 
While it may seem overly cautious or unrealistic at this stage of wearable technology, we want to maintain our effort to reduce the exposure of our volunteers and ourselves to signals that have not yet proven to be risk-free.

In the present work, we focus on passive solutions such as the one presented in \cite{Expressure}, where a textile pressure matrix (mechanomyography) was introduced into a headband design to classify seven facial expressions, with intermediate results of up to 38.00\% accuracy. 
In \cite{bello2020facial} acoustic mechanomyography (AMMG), another passive technique was employed with an F1 score of 54.00\% in classifying ten facial muscle activities. 
Piezoelectric thin films (PEF) have been used in real-time \cite{PiezoFacialNature} to detect and classify skin deformation to decode facial movements in patients with amyotrophic lateral sclerosis.
Their work is intended to be used in clinical settings for nonverbal communication and neuromuscular monitoring conditions. 
PEF sensing technology is lightweight, customized, and with mechanical harvesting capability \cite{PiezoHarvesting}; therefore, we could claim that PEF technology is worthy of research and study in specific applications.  
Here, we proposed to fuse passive sensing such as; pressure mechanomyography (PMMG) using a force-sensitive resistor (FSR) and piezoelectric film (PEF), inertial sensing based on orientation and acceleration, and acoustic mechanomyography (AMMG). 
\vspace{-10pt}

\subsection{Mechanomyography (MMG)}
Cyclic patterns of muscle contraction and relaxation are common in many everyday human activities, such as cycling, running, and laughing, to name a few.
How the muscle activity pattern changes in intensity, duration, and frequency to understand human behavior is the topic of many research works. 
Electromyography (EMG) is a dominant option for measuring muscle activity in a laboratory setting; the standard procedure is to use surface or needle electrodes embedded in the muscle. 
EMG has limitations in terms of cost, is often restricted to a controlled environment, and signal intensity is susceptible to skin impedance, age, and weight of the individual \cite{MMGIMUNature,EMGLimitations}. 
In \cite{MMGIMUNature}, the authors compared EMG with a fusion based on MMG and acceleration data for automatic segmentation and concluded that the results of the MMG+acceleration combination are similar to those of EMG, making the MMG option a suitable solution for natural settings such as the home.

\textbf{Mechanomyography (MMG)} is also suitable for assessing skeletal muscle activities and monitoring fatigue and force imparted in muscle movement \cite{MMGAndMuscleReview, MMGAccelFatige}.
In the literature, the measurement of MMG  typically employs IMU \cite{MMGAccelFatige}, piezoelectric \cite{MMGAccelParkinsons}, microphones, pressure sensors, laser distance sensor, and others \cite{MMGSensorChoice}. 
Multimodal MMG solutions have been studied, as in \cite{MMGAccelParkinsons}; The authors combined forearm acceleration and piezoelectric MMG sensors in the biceps brachii, brachioradialis, and pectoralis major to study mechanical muscle oscillations in patients with Parkinson's condition. 
In \cite{EMGMMG}, a myoelectric fusion (MMG+EMG) is proposed to overcome the unreliable sensor-skin interface in upper limb prosthesis manipulation.
It follows from all the above that MMG is a broad and active area of research.
In our work, there is particular interest in two categories of MMG; acoustic mechanomyography (AMMG) and pressure mechanomyography (PMMG).

\textbf{Acoustic mechanomyography (AMMG)} is based on monitoring muscle force (contraction/relaxation) by using low-frequency signals (2-200 Hz) and power signals below 50 Hz  \cite{MMGFusion}.
Using a microphone on the participants' biceps brachii muscle during lifting activities in \cite{MechanomyographyFatigue} shows the AMMG as an alternative to the electromyogram for measuring muscle fatigue. 
In \cite{MicVsImuMotionArtifact}, the authors found that AMMG was significantly less influenced by motion artifact than the corresponding accelerometer spectra ($ p \leq 0.05$). 
They conclude that condenser microphones are preferred for MMG recordings when mitigating motion artifact effects is essential.
This is especially true for kinesiological studies involving limb movement, such as cycling. 
AMMG for facial muscle movement recognition was introduced in \cite{bello2020facial}.
The stethoscope-based design in \cite{bello2020facial} amplifies the intensity of muscle movement sounds at the expense of user comfort due to the large diameter. 
We propose to sense AMMG with two Inter-IC Sound (I2S) digital microphones on the frontalis muscle. 
I2S microphones are small compared to Electrec microphones in \cite{bello2020facial}, and because they are digital microphones, their output is less disturbed by subtle movements of the wearable accessory.

\textbf{Pressure mechanomyography's (PMMG)} prime purpose is to transduce muscle vibration changes by means of pressure patterns.  
PMMG has been used for facial expression and cognitive load in \cite{Expressure}. 
The authors have employed a textile pressure matrix (TPM) on the forehead of twenty volunteers to recognize seven facial expressions from the Warsaw photoset \cite{Warsaw}. 
The 38.00\% (14.00\% chance-level) accuracy in recognizing seven facial muscle activities with an unobtrusive, textile, and passive sensing method shows encouraging results.        
\section{InMyFace Approach}
\label{sec:approach}
\begin{figure*}[!t]
    \centering
    \includegraphics[width= 0.9\textwidth]{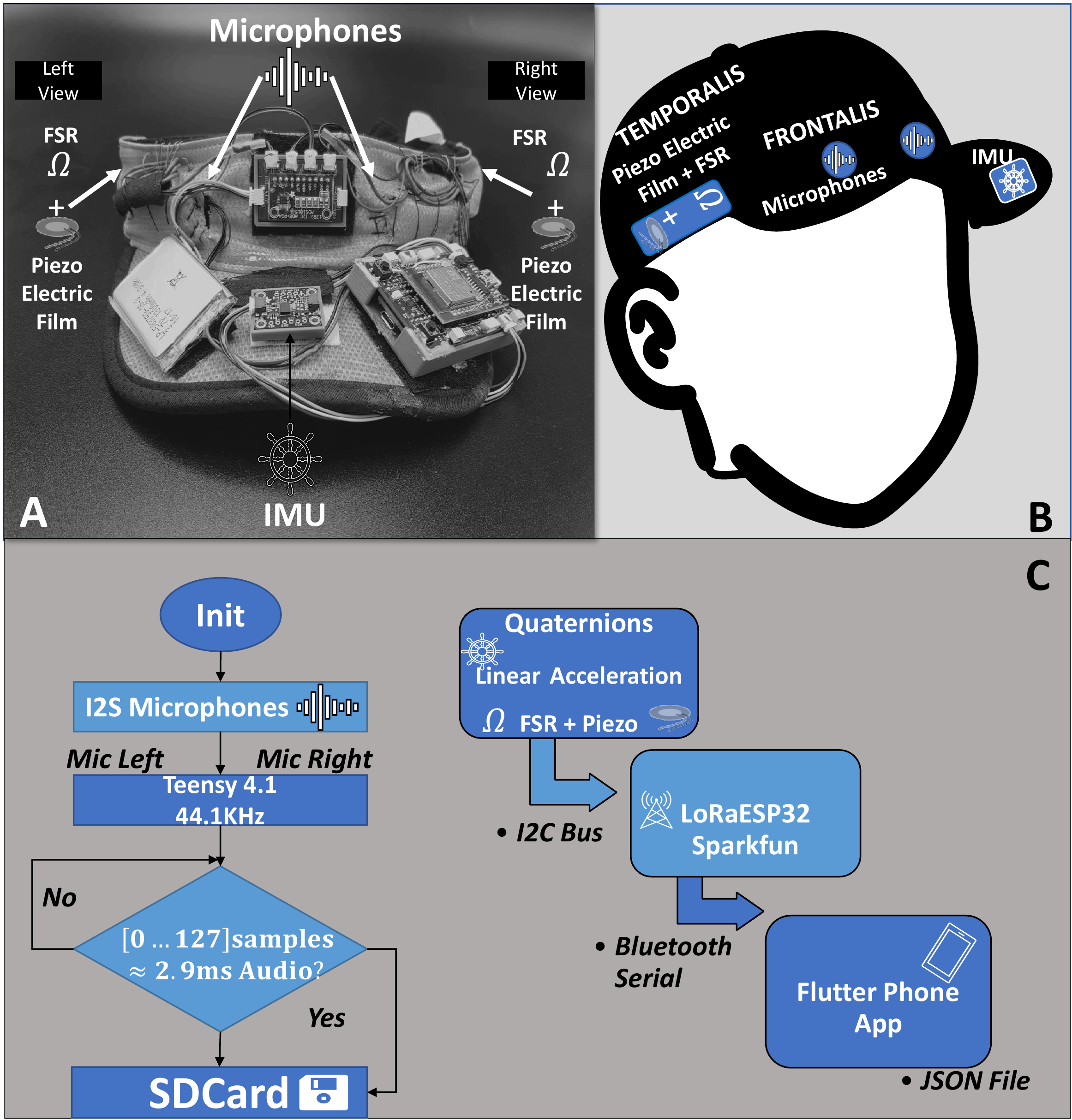}
    \caption{Hardware Prototype and Data Collection Diagram. \textbf{(A)} Sports Cap with Sensors Distribution. \textbf{B} Sensor Placement on the Frontalis and Temporalis Muscles of the Participant. \textbf{(C)} Data Acquisition Diagrams with the Custom Printed Circuit Board.}
    \label{fig:Sportcap}
\end{figure*}
Our system combines inertial, pressure, and audio sensors to recognize facial muscle activity from an unobtrusive sports cap platform. 
\cref{fig:FacialSignals} presents the facial muscle movements dictionary. 
The IMU has already demonstrated its potential to distinguish various face- and head-related movements \cite{FaceCommandGlasses, KissglassIMUOculo}. 
PMMG provides a flexible and comfortable solution for facial gesture recognition with moderate accuracy \cite{Expressure}.  
AMMG for facial muscle activity recognition was used in \cite{bello2020facial}. 
However, the design was bulky and obtrusive to achieve high sensing accuracy. 
Combining the three sensing modalities with an appropriate sensor fusion pipeline allows us to achieve high accuracy with an unobtrusive system.

\subsection{System Architecture and Implementation}
The prototype hardware is shown in \cref{fig:Sportcap}.   
The FSR sensors were distributed on the frontalis and temporalis muscles using a sports cap as a wearable accessory. 
Two Inter-IC Sound (I2S) microphones sampled at 44.1kHz were placed on the left/right side of the cap as shown in \cref{fig:Sportcap}. 
A custom printed circuit board (PCB) based on Teensy 4.1 \footnote{Teensy 4.1, Paul Stoffregen: https://www.pjrc.com/store/teensy41.html DLA: February 07, 2023} and LoRaESP32 \footnote{LoraESP32, Sparkfun: https://www.sparkfun.com/products/18074  DLA: February 07, 2023}  was used to sample the sensor data.
The integrated SD card of the Teensy 4.1 is used to store the acoustic data. 
The microphone openings are pointed towards the frontalis muscle to capture the mechanical sound information (AMMG).
The analyzed sound information comes from the muscle movements, so it is a privacy-aware system (no speech or ambient sounds). 
Pairs of FSR and PEF were placed to the left and right of the temporalis muscle. 
The FSR and PEF were selected to measure the PMMG generated by facial muscle movements. 
An IMU was in the sports cap viewer to capture head movements related to facial expressions. 
FSR, PEF, and IMU data were transferred by Bluetooth Serial (BT) to a cell phone application (Flutter Framework \cite{Flutter}) to save them in a JSON file for further analysis.
The sampling rate was about 100 Hz for FSR, PEF, and IMU data. 
The Bluetooth communication scheme limited the data acquisition but is still fast enough to capture micro and macro expressions with a duration $\leq 200 ms$ and duration $\geq$ 200 ms, respectively \cite{MicroMacroExp}. 
Detailed sensor data acquisition diagrams are in \cref{fig:Sportcap}.

The sensor distribution (around the head) and the dimensions/weights of the selected sensors, see \cref{table:Sensors}, make our design suitable for integration into other head-related accessories such as headbands and glasses. 
In our sports cap, the dimensions and weight are negatively affected by the selected battery (1300 mAh) and the main board of the prototype (6.5 x 4.3 cm); both could be reduced in a future design and improve user comfort.

The power consumption of the prototype is around 0.22A at 4.97V (1.09 Watts continuous mode) \footnote{USB Digital Power Meter: \url{https://www.az-delivery.de/en/products/charger-doktor}  DLA: February 07, 2023}.  
A number of possibilities exist for further power consumption reduction based on the overall concept. 
On the one hand, FSR-PEF data could trigger IMU and audio data acquisition, reducing power consumption to 0.06A at 4.99V (0.3 Watts) when no pressure is detected in the temporalis muscle.  
On the other hand, PEF is a mechanical energy source.
Mechanical energy is considered ubiquitous ambient energy that can be converted into electric power \cite{PiezoHarvesting}. 
Employing piezoelectric as a mechanical energy harvesting mechanism is an active field of research \cite{PiezoHarvesting, ShoePiezo, Wristwoodpecker}. 
Harvesting energy from human motion and at the same time classifying such motions has also been demonstrated before \cite{PiezoFromBody, PiezoTextil, ShoePiezo, PiezpWearableUpperLimb} which could be employed to reduce the power consumption further.

In addition to mitigating power issues, the FSR has the advantage of being robust against motion artifacts.
We thus recommend the use of the slope/gradient of the FSR signal as a possible trigger for automatic segmentation of the input data and to avoid motion artifacts.  
For example, the signal's slope of acceleration data was automatically used to segment MMG data in \cite{MMGIMUNature}.
Finally, although we use off-the-shelf and non-textile FSR and PEF sensors for fast prototyping, it is noteworthy that both technologies are already available in textiles \cite{Expressure, PiezoTextil, PiezoMaterialFlex}.

\begin{table*}[!t]
\caption{Sensors Characteristics} 
\renewcommand{\arraystretch}{1}
\resizebox{\textwidth}{!}{
\setlength{\tabcolsep}{3pt}
\begin{tabular}{m{10em} m{15em} m{10em} m{3em} m{15em}}
\hline
Sensor & Manufacturer Name & Dimensions (cm) & Weight (grams) & Benefits\\
\hline 
FSR $^1$
& Alpha MF01A-N-221-A01 
& 1.25 diameter 
& 0.26 
& Ultra-thin and flexible \\
PEF $^2$
& TE SDT1-028K shielded
&  4.45 x 1.97 x 0.32 
& 0.30 
& Low noise, shielded and flexible\\
Microphones $^3$
& Knowles SPH0645LM4H 
&  0.35 x 0.26 x 0.09 
& 0.40
& High SNR of 65dB(A), Flat Frequency Response, Omnidirectional\\
IMU $^4$
& Bosch BNO055
&  0.38 x 0.52 x 0.11 
& 0.15
& Outputs fused sensor data\\
\hline
\label{table:Sensors}
\end{tabular}}
\raggedright
\scriptsize{$^1$ \url{https://www.mouser.de/datasheet/2/13/MF01A__c3_a2_c2_96_c2_a1_A01-1915118.pdf} DLA: February 07, 2023 \\}
\scriptsize{$^2$ \url{https://www.te.com/usa-en/product-CAT-PFS0010.html} DLA: February 07, 2023 \\}
\scriptsize{$^3$ \url{https://media.digikey.com/pdf/Data\%20Sheets/Knowles\%20Acoustics\%20PDFs/SPH0645LM4H-B.pdf} DLA : February 07, 2023\\ }
\scriptsize{$^4$ \url{https://www.mouser.de/datasheet/2/783/BST_BNO055_DS000-1509603.pdf} DLA: February 07, 2023}
\vspace{-10pt}
\end{table*}

\subsection{Multimodal Sensor Fusion}
\begin{figure*}[!t]
     \centering
     \begin{subfigure}[b]{0.49\textwidth}
         \centering
         \includegraphics[width=\textwidth]{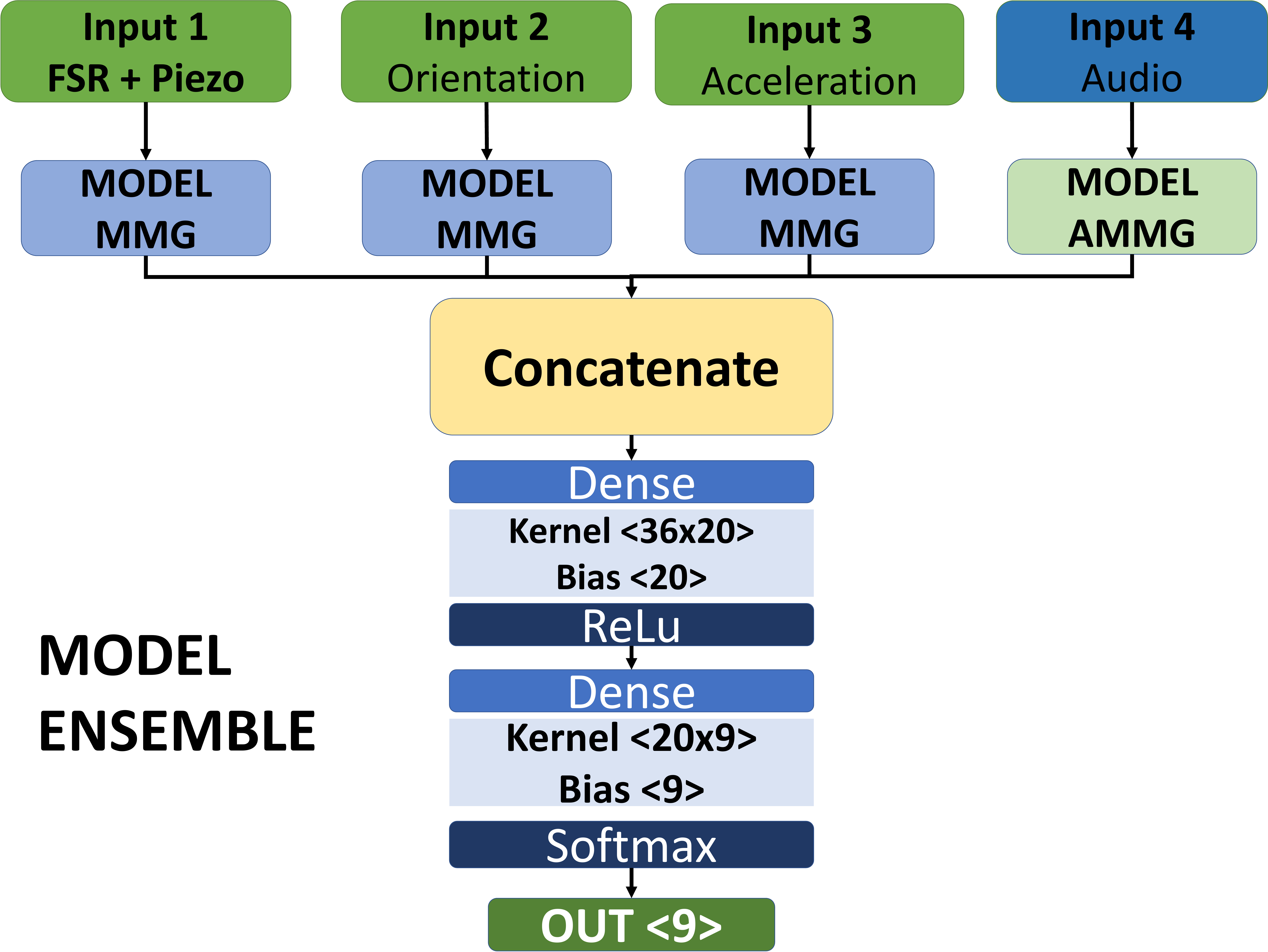}
         \caption{\textbf{Left}}
         \label{fig:NNEnsemble}
     \end{subfigure}
     \hspace{-5pt}
     \begin{subfigure}[b]{0.49\textwidth}
         \centering
         \includegraphics[width=\textwidth]{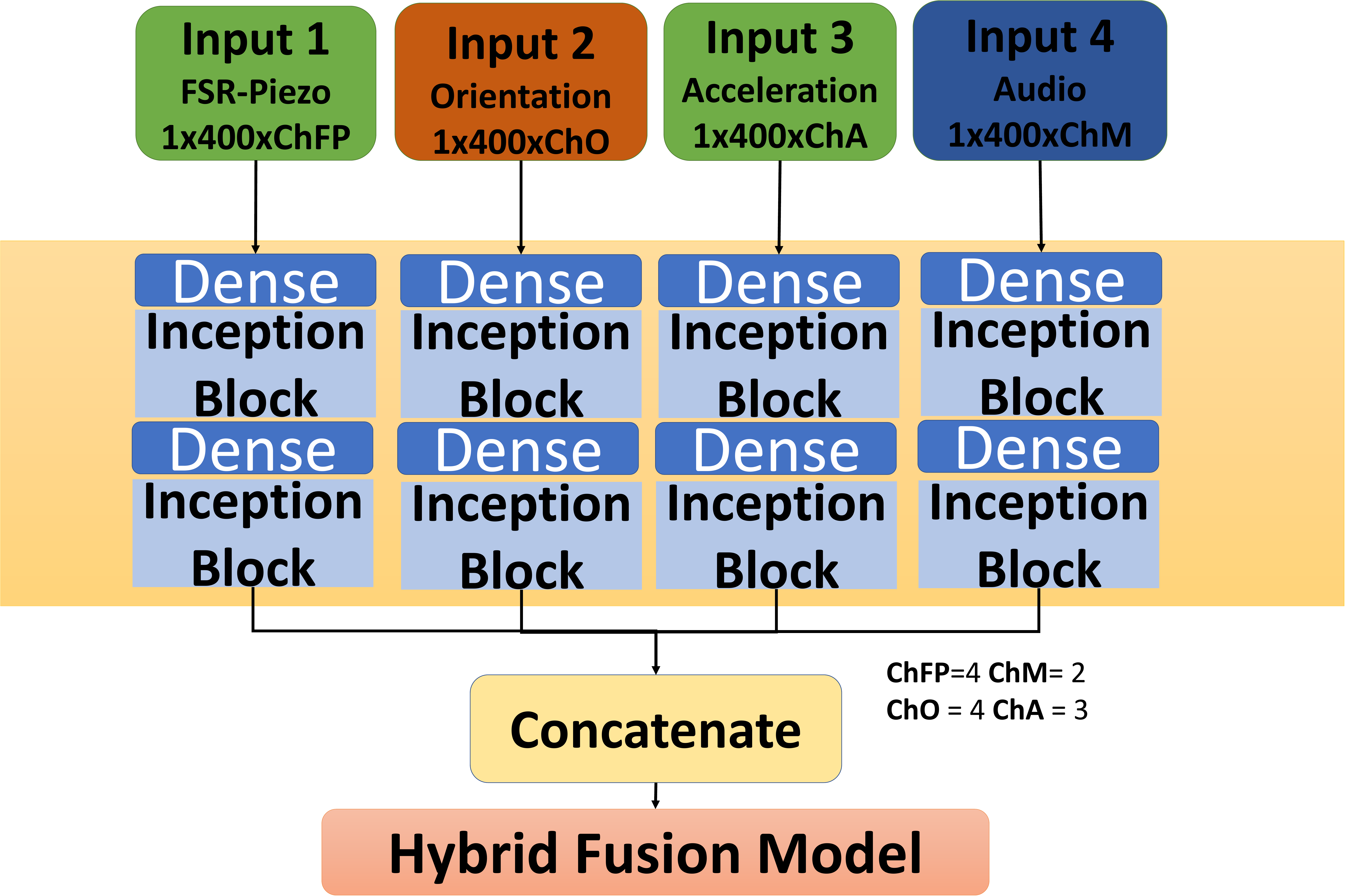}
         \caption{\textbf{Right}}
          \label{fig:NNHybridFusion}
     \end{subfigure}
     \caption{(a) \textbf{Left} Ensemble Multimodal Sensor Fusion Model Overview; Fusion in the Prediction Phase. (b) \textbf{Right} Hybrid Multimodal Sensor Fusion Model Overview; Fusion Within Hidden Layers, and Before Prediction.}
     \label{fig:MultiFusion}
\end{figure*}

Multimodal sensor fusion can be broadly classified into early and late fusion, depending on the position of the fusion within the processing chain \cite{EarlyVsLate}.  
Our primary approach is referred to as late fusion. 
The fusion is performed in the individually trained networks' decision phase (confidence scores). 
The late fusion method has the advantage of extracting the specific patterns of each sensor independently and the parallel deployment of each sensor-dependent NN on multiple MCUs, reducing the recognition latency and memory requirement per MCU($\leq$ 2 MB per network). 
The main drawback of late fusion is the limited potential to extract cross-correlation between sensing modalities and channels. 
Additionally, we also explored a hybrid fusion alternative.
The fusion performed in the hidden layers of the neural network and before the decision layer is called hybrid fusion. 
In our work, the hybrid fusion structure and evaluation are made considering the outcomes from the late fusion performance. 
An overview of the late fusion and hybrid fusion diagrams is depicted in \cref{fig:MultiFusion}. 
The \cref{fig:NNEnsemble} \textbf{Left} shows the concatenation of the sensor-dependent models after the decision phase using an ensemble NN.
The specific sensor-dependent models are explained in \cref{sub:MultiEnsemble}. 
The \cref{fig:NNHybridFusion} \textbf{Right} presents the primary blocks of the hybrid fusion method. 
The blocks consist of sensor inputs, inception blocks, and hybrid fusion NN. 
The details of each block are described in \cref{sub:MultiHybrid}.

\subsubsection{Multimodal Ensemble Late Sensor Fusion} \label{sub:MultiEnsemble}
\begin{figure*}[!t]
    \includegraphics[width= \textwidth]{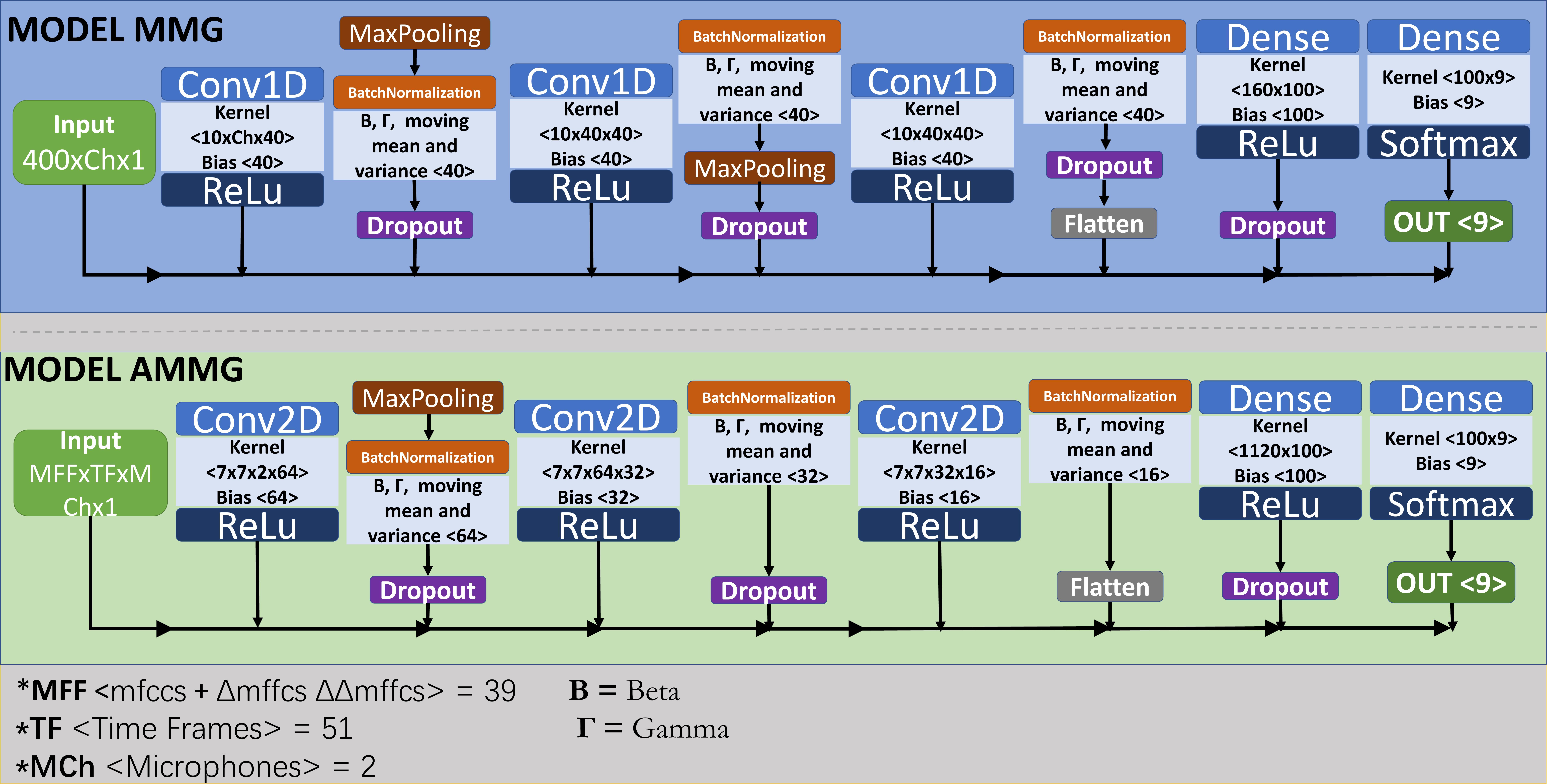}
    \caption{Sensor Dependent Neural Network Models. \textbf{(Model MMG Top)} Neural Network for FSR-PEF (PMMG) and IMU(Orientation and Acceleration) Information. \textbf{(Bottom Model AMMG)} Neural Network for Mel Frequency Cepstral Coefficients of Audio (AMMG).}
    \label{fig:NNIndSensors}
\end{figure*}

We employ sensor-based late fusion \cite{CNNMultiFusion} as depicted in \cref{fig:NNEnsemble} \textbf{Left} and \cref{fig:NNIndSensors}. 
The data was divided by sensor type, and we implemented four sensor-dependent neural networks (NN) models in \cref{fig:NNIndSensors}. 
This approach gives each NN the advantage of learning the unique properties of each modality and facilitates a simple fusion method.  
In the initial step, the inputs to the sensor-dependent NNs were pre-processed in a sensor-specific manner, as described below. 
The data of each movement was processed as a whole instance. 
The NNs were developed using the TensorFlow framework (version 2.9.2).
The training included early stopping with the patience equal to 30 and restored weights option equal to true to avoid overfitting and ran for 500 epochs.
In the individual models, the learning rate was manually tuned for each participant, and for the case of one ensemble model for all participants (cross-user case), it was set to 0.03.
Categorical cross-entropy loss function and Adagrad optimizer were used to optimize the sensor-type NN.
The ensemble NN consisted of one fully connected layer of 20, Adam optimizer (0.01), and softmax function with nine probability outputs, see \cref{fig:NNEnsemble} \textbf{Left Model Ensemble}. 
The details of the sensor-type dependent NNs are explained below.

\textbf{Pressure Mechanomyography (PMMG) and IMU:}
We sense PMMG with a combination of FSR and PEF. 
For the case of inertial sensing, the quaternions were selected for orientation to avoid the gimbal lock problem \cite{Gimbal} and to improve stability. 
The FSR, PEF, and inertial sensors (quaternions and acceleration) data were normalized by subtracting the average of the gesture's first (starting point) and last values (ending point).
Since each facial event is a temporal series with variable lengths, a dynamic resample procedure to 400 samples was applied \cite{bello2021mocapaci, bellomove}.
Then, the resampled signals were fed to a first-degree Butterworth low pass filter with a cut frequency of 5Hz to remove the ringing peak in the signal's edges coming from the resample procedure and to highlight the low-frequency range.

FSR and PEF signals were treated as a pair; hence they have a joint NN. 
Orientation and acceleration were processed in separate NNs. 
In total, three networks were trained for PMMG and inertial sensing. 
The NN structure was based on a modified 1D-LeNet5 network \cite{LeNet5} as depicted in \cref{fig:NNIndSensors} \textbf{Model MMG}.
The network consisted of a convolution (conv)—max pooling (maxpool)-conv-maxpool-conv—fully connected (fc)-fc-softmax layers with batch normalization and dropout on the convolution layers.
The convolution layers contain 40 filters, a kernel size of 10, and the activation function ReLu. 
For max pooling, the pool size was (40, 40) for the first convolution (400, 40) and (4, 40) for the second convolution (40, 40).
The third convolution was of size (4, 40) without pooling. 
A flattening layer of 160 was followed by a fully connected layer of 100.
The nine outputs for the different facial muscle activities in \cref{fig:FacialSignals} are then converted into probabilities by a fully connected layer and softmax function. 
A detailed view of the sensor-dependent neural network structure for the PMMG and IMU is shown in \cref{fig:NNIndSensors} \textbf{Top Model MMG}.

\textbf{Acoustic Mechanomyography (AMMG):}
As shown in \cref{fig:Sportcap}, two I2S microphones sampled at 44.1kHz were positioned on the sports cap to cover the frontalis muscle of the volunteer.   
The audio information was used as AMMG as proposed in \cite{bello2020facial} but without the stethoscope to amplify the audio. 
Two channels of audio were resampled to 52000 (1.17 seconds). 
Muscle's audio data was transformed to the mel spectrum to reduce the dimension of the audio and speed up the convergence time of the neural network for a small dataset. 
The Mel Frequency Cepstral Coefficients (MFCCS) have been used for many applications \cite{PPGandMel,MelSpeech2}. 
Thirteen Mel filters are a common choice in the literature for audio analysis \cite{MelSpeech,MelSpeech2}.
The short-time Fourier transform (STFT) parameters were defined as; a sampling rate of 44.1kHz,  hamming window and size of 4096 (93 ms), and hop length of 1024 (23 ms). 
Although we are not using speech information, the STFT configuration was selected to match the default configuration defined in \cite{Librosa} (the typical setting for speech analysis) for simplicity. 
In addition to the MFCCS, we also calculated the first and second derivatives of the MFCCS to boost the recognition accuracy \cite{MFCCsDelta}. 
The input shape to the NN was (39,51,2); 13 MFCCS, 13 MFCCS' delta, and 13 MFCCS' second delta, 51-time frames, and two audio channels. 
The deep learning model was defined as two-dimensional convolution (conv2D)-max pool-conv2D-conv2D-fc-fc-softmax layers with batch normalization and dropout on the convolution layers. 
The first, second, and third conv2D consisted of 64,32, and 16 filters, respectively, with a kernel size of 7 and an activation layer of ReLu as depicted in \cref{fig:NNIndSensors} \textbf{Bottom Model AMMG}.

In the case of a future real-time implementation, the modular approach offers the flexibility of sharing the computation of each NN model between microcontrollers (MCUs). 
The NNs developed in this work are less than 2 MB, a practical size for an embedded device (typically between 4 and 16 MB of flash), and with the TensorFlow Lite framework\footnote{Machine Learning for Mobile and Edge Devices - TensorFlow Lite \url{https://www.tensorflow.org/lite}}, can be compressed for use on mobile/embedded devices. 
Compared to a wearable camera solution with 162 MB in \cite{Neckface}, our 2 MB distributed NNs are two orders of magnitude smaller. 
Therefore, we could schedule the FSR-PEF and IMU model (\cref{fig:NNIndSensors} \textbf{Model MMG}) into MCU-one and the audio model ( \cref{fig:NNIndSensors} \textbf{Model AMMG}) into MCU-two, which in the end will be merged by the ensemble model using the Bluetooth-capable phone. 
As lightweight and individually deployable models in the MCU(s), only the prediction results are sent out of the embedded device, which will maintain the privacy of the user's data.

\subsubsection{Multimodal Hybrid Fusion}
\label{sub:MultiHybrid}
\begin{figure*}[!t]
\includegraphics[width=\textwidth]{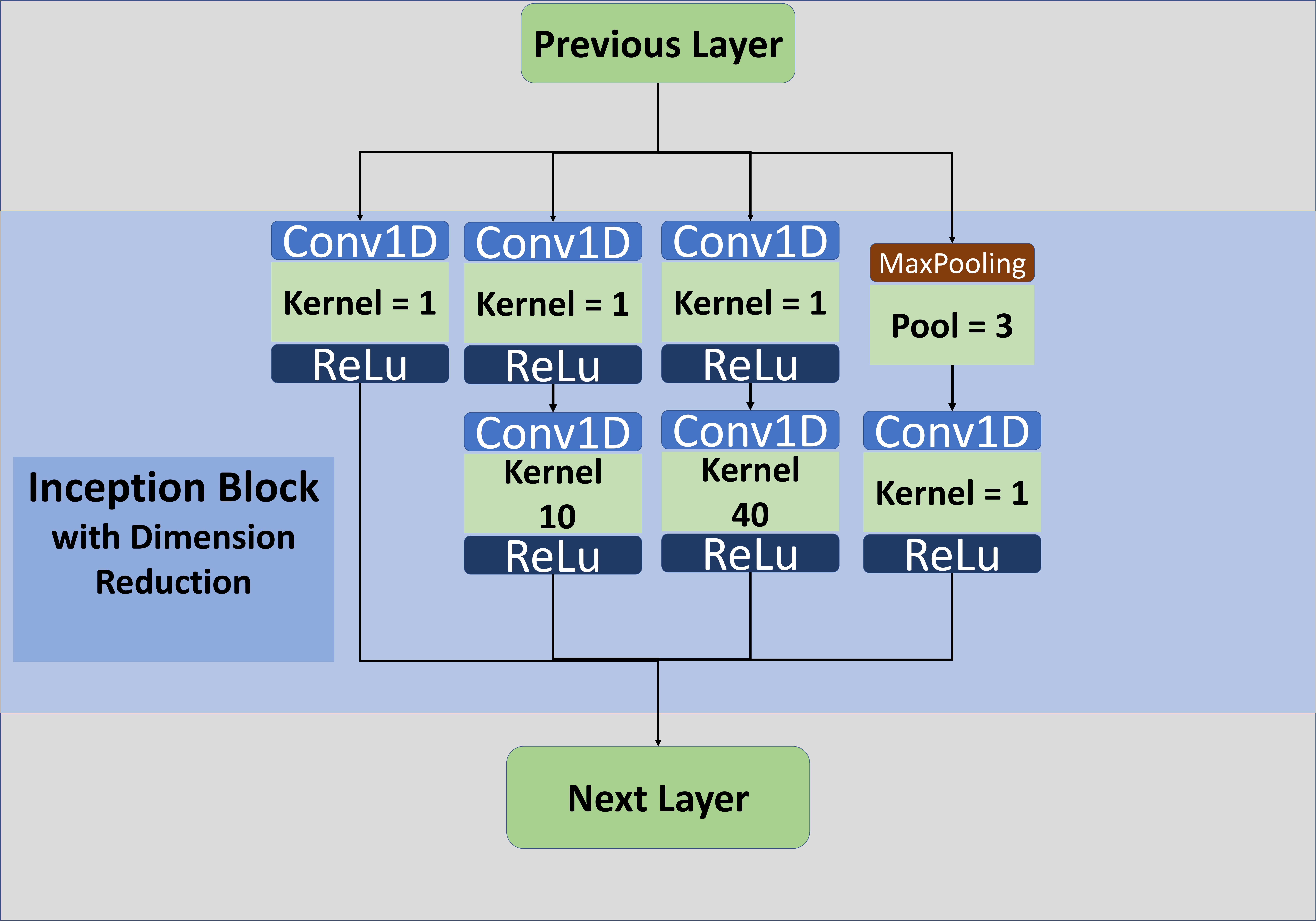}
    \caption{Modified Inception Block with Dimension Reduction Locally Connected per Sensing Modality}
    \label{fig:InceptionBlock}
\end{figure*}
\begin{figure*}[!t]
\includegraphics[width=\textwidth]{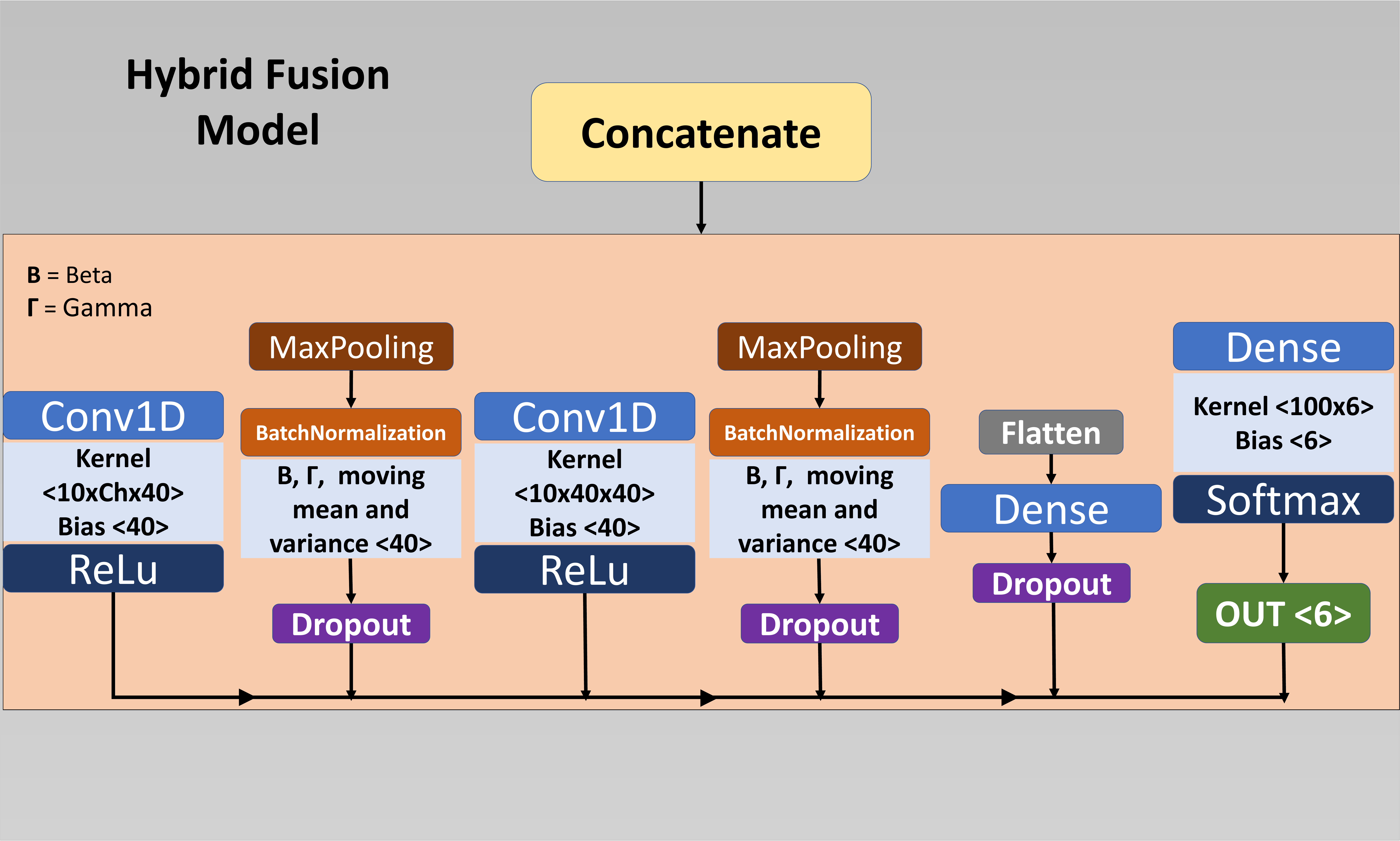}
    \caption{Multimodal Hybrid Fusion Model}
    \label{fig:HybridModel}
\end{figure*}
An early/hybrid fusion approach learns the contributions to the recognition performance of all sensing modalities as a unit, exploiting the cross-correlation between information sources. 
On the other hand, the NN structure for early/hybrid multimodal sensor fusion could be complex to give competitive results, consequently increasing computational time \cite{CNNMultiFusion,EarlyVsLate}.   
In some cases, fusing low-level features might be irrelevant to the task, thus decreasing the fusion power \cite{ThermalColorEarlyLateComp}. 
Hence, the level at which the merge is made will influence the quality of the information being fused. 
Additionally, this fusion method requires all the sensor's data to enter the same processing pipeline, reducing the parallelism and increasing deployment complexity in embedded devices.
The challenge is maintaining a low-complexity model while achieving performance comparable to the late fusion approach. 
In this section, we evaluate the performance of such a sensor fusion methodology to test our system's facial muscle movements recognition capability.  

Our experiment is based on imitating facial movements related to typical facial expressions. 
Therefore, the ground truth could be affected negatively by the imitation ability of the volunteers (with no acting experience).
In this section, the eight best imitators were selected to test the system's feasibility without outliers due to the imitation inaccuracy of our participants. 
The selection is based on the visual perception of the similarity of the volunteers' facial dictionary imitation performance, as assessed visually by most participants. 
The main drawback of reducing the number of volunteers is that it increases the risk of overfitting, which is already high due to the higher complexity of hybrid fusion modeling compared to late fusion for our specific task.
Therefore, we employ techniques developed to reduce the risk of overfitting. 
These techniques include: early stopping with patient equal to 30 and restored weights option enabled, batch normalization, and max pooling, the evaluation scheme is defined by leaving one session out, and the NN structure is based on a lightweight NN designed to reduce the number of parameters. 
We focused on using the main block of one of the most popular lightweight NN as the stem network \cite{Inception}, locally connected to each sensing modality, known as the inception block as shown in \cref{fig:InceptionBlock}. 
These main blocks were explicitly designed to reduce the number of parameters while maintaining a balance between latency and accuracy to allow their deployment in mobile/embedded systems. 
Moreover, the loss function performs soft labeling (also called label smoothing) with $10\%$ distributed equally over the opposite classes. 
Hard labeling is the typical way of assigning labels to class members, where class membership is binary, i.e., labels are true/false.
In soft labeling, class membership is based on a probability score assigned to each member.
Thus, a 10 \% soft labeling means that, for example, the probability of class 1 is 90 \% when class 1 is the truth value, and 10 \% is equally distributed over the opposite classes.

In order to reduce the complexity of the hybrid model structure, we have reduced the number of classes based on the risk of confusion. 
The joy and surprise classes are merged due to the similarity of visual perception within facial gestures.
These two classes differ only in the mouth movement, and the placement of the sensors on the forehead makes their recognition indirect, which could lead to considerable confusion between them. 
In addition, the following section \cref{sec:results} shows that the confusion in the late fusion approach is mainly dominated by two pairs of classes, "Disgust-Anger" and "Sadness-Anger", when all sensing modalities are combined, leading us to merge these categories. 
The classes recognized were Neutral, Joy + Surprise, Disgust + Sadness + Disgust, Wink, Fear, and Take a pill, for a total of six categories. 
Despite all the parameters and complexity reduction techniques, the hybrid fusion in our case reached 939,706 parameters ( 2.35x Late Fusion Parameters). 
However, based on the number of parameters is still a small network compared to typical networks considered smallish, such as MobileNetV2 with 3.5 million parameters \cite{mobilenetv2}.

As depicted in \cref{fig:NNHybridFusion}, we employed inception blocks locally connected to each sensing modality, followed by a concatenation. The output is then fed to a hybrid fusion model/NN. 
Before entering the NN, the same signal processing steps performed in \cref{sub:MultiEnsemble} are applied per sensor type.
Our data sources are not spatiotemporally aligned due to the number of channels/resolutions and different sampling frequencies. 
Therefore, a certain degree of preprocessing is necessary before concatenating them using a CNN.  
As a result, the data of all sensors were in the form (Time steps $=$ 400, Channels). 
The channels were four, four, three, and two; for PEF and FSR, orientation, linear acceleration, and audio, respectively. 
The locally connected inception block structure is depicted in \cref{fig:InceptionBlock}. 
The inception block is then followed by a hybrid fusion model as shown in \cref{fig:HybridModel}. 
The hybrid model consisted of conv1D-maxpool-conv1D-maxpool-flatten-fully connected(fc)-(fc)-softmax layers with batch normalization and dropout on the convolution layers. 
The convolution layers contain 40 filters, a kernel size of 10, and the activation function ReLu. 
For max pooling, the pool size was (40, 40) for the first convolution (400, 40) and (4, 40) for the second convolution (40, 40).
A flattening layer of 160 was followed by a fully connected layer of 100.
The six outputs for the six categories (Neutral, Joy + Surprise, Anger + Disgust + Sadness, Winking, Fear, and Taking a Pill) are then converted into probabilities by a fully connected layer and softmax function. 
The training ran for 200 epochs with early stopping enabled with a patient equal to 30 and restoring weights option enabled.
Categorical cross-entropy loss function and AdaDelta optimizer with a learning rate of 0.09  were used to optimize the sensor-type NN.
AdaDelta optimizer is a method that performs an adaptive learning rate per dimension, and its main advantage is that there is no need to select a global learning rate. 
Moreover, it can handle the intrinsic continuous decay of learning rates throughout training \cite{AdaDelta}. 
A summary diagram of the techniques used to build the hybrid neural network structure and training is shown in \cref{fig:ResultsHybrid} \textbf{Right}.

\section{Results and Discussion}
\label{sec:results}
\subsection{Experimental Procedure}
Thirteen participants from diverse backgrounds (Germany, Italy, Peru, India, France, China, Republic of Korea, and Venezuela) mimicked the facial muscle movements defined in \cref{fig:FacialSignals} in a random sequence per session while wearing our sports cap prototype.
The dictionary in \cref{fig:FacialSignals} contains seven of the facial expressions proposed in \cite{Warsaw} plus two additional gestures from \cite{bello2020facial}. 
It is important to emphasize that the system recognizes consciously simulated facial expressions and not authentic expressions, as in previous hardware-related works (see Table \ref{table:comparison}), to test the feasibility of fusing the sensing modalities. 
Therefore, we emphasize that what we recognize are  "facial muscle movements" related to creating features similar to what we expect to see in genuine facial expressions. 

Participants' muscle movement sounds and pressure patterns were recorded without any additional conditions other than that they mimic the dictionary as closely as possible.
Subjects were not forced to make sounds or restricted to a specific time to perform the facial muscle movements. 
Therefore, the amount of time for each gesture is variable, even for the same participant. 
All participants performed five sessions.
One session consisted of four randomized/shuffled appearances of each face gesture within the dictionary, which is used to avoid muscle fatigue and avoid correlation between instances of the same gesture. 
A total of 180 instances per volunteer were collected. 
The neutral face marks the start and end points of a gesture. 
In total, 2160 valid facial movements were collected. 

Mechanomyography provides the timing requirements for the experiment to avoid corrupting the data because of the tiredness of the participant \cite{MechanomyographyFatigue}. 
Therefore, the duration of a session was between five and seven minutes.
On average, subjects rested for at least 10 minutes between sessions (without wearing the sports cap). 
For some volunteers, the experiment was completed in 2 days. 
The volunteers were six women aged 21 to 30 years and seven men aged 21 to 35 years (mean $27.00\pm4.11$) with head diameters of 52 to 61 cm (mean $55.54\pm2.56$) and with different hairstyles (straight, curly, and no hair). 
The experiments were carried out in an office, and the participant remained seated during the experiment.
All participants signed an agreement following the policies of the university's committee for protecting human subjects and following the Declaration of Helsinki \cite{Helsinki1975}. 
The experiment was video recorded for a further confidential analysis. 
The observer and participant followed an ethical/hygienic protocol following the mandatory public health guidelines at the date of the experiment.
The minimum number of valid sessions was 4 (Volunteers 3,4,8, and 10), and the typical case was five valid sessions.
The training/testing scheme was defined as 5-fold stratified cross-validation with leave one session out scheme \cite{StratifiedCrossVal}, similar to the cross-session validation in \cite{EarIO}. 
Leave one session out cross-validation reinforces the robustness of the training against re-wearing of the system, as is common with wearable devices.

\subsection{Results}
\begin{figure*}[!t]
    \includegraphics[width=\textwidth]{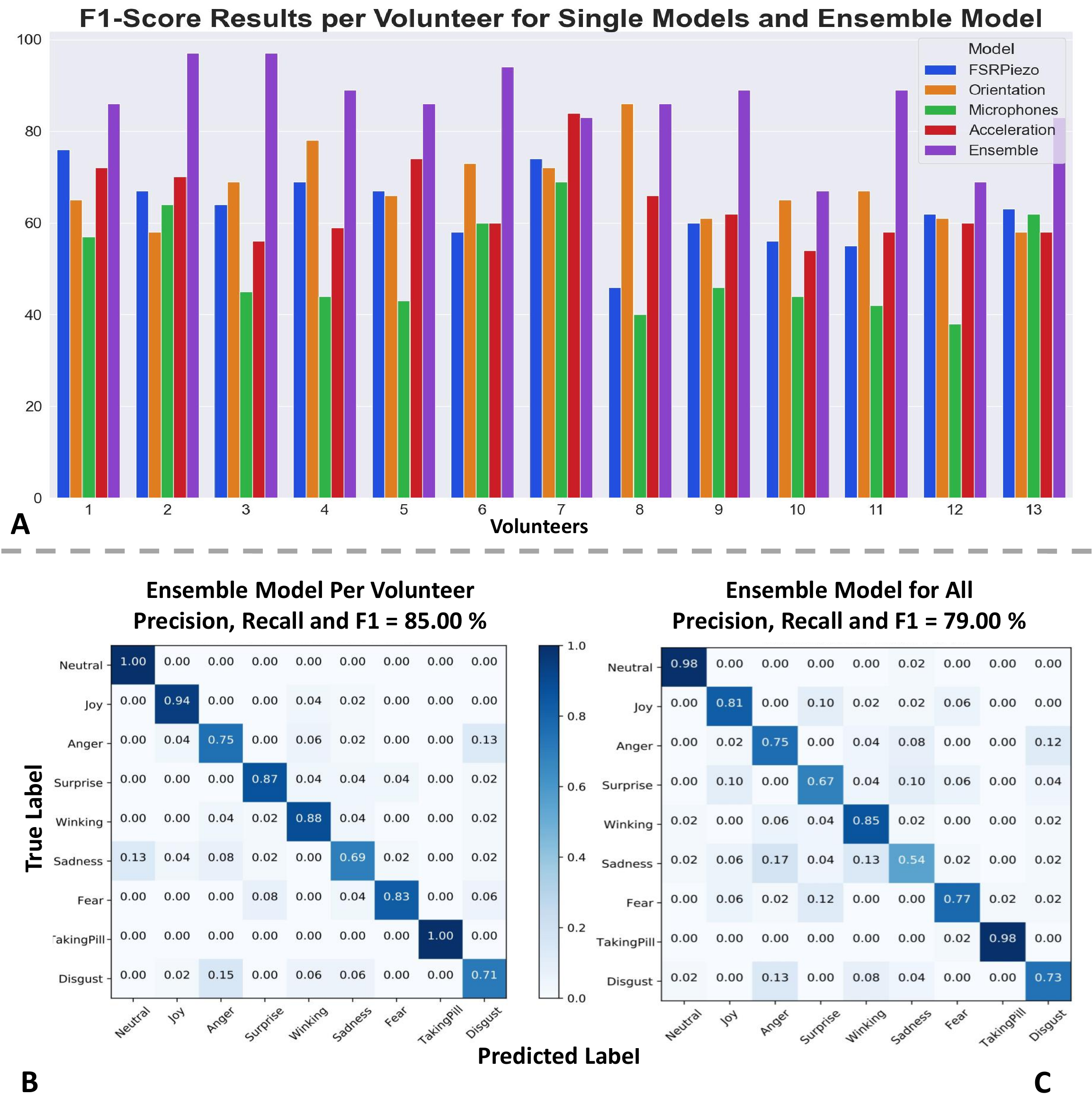}
    \caption{Results Sensor Dependent and Ensemble Neural Networks \textbf{(A)} F1-Score of Sensor Dependent Neural Network per Volunteer Comparison. \textbf{(B)} Recall Results of Ensemble Model per Participant Avg F1=85.00\%. \textbf{(C)} Recall Results One Ensemble Model for All Participants F1=79.00\%}
    \label{fig:Results}
    \vspace{-15pt}
\end{figure*}

\cref{fig:Results} \textbf{A} shows the performance in the case of individual models (per-user) versus the ensemble model. 
The average improvement of the ensemble model was 14.30\% (F1 score) compared to the best sensor-dependent model. 
The confusion matrix in \cref{fig:Results} \textbf{B} shows an average F1 score of 85 \% for the case of the individual concatenated models.
The ensemble model for all (cross-user, see \cref{fig:Results} \textbf{C}) yielded an F1-score of 79.00\%.
In addition, the one-ensemble model for thirteen participants achieved a 16.00\% increment in F1-score compared with the best sensor-type dependent NN (see \cref{fig:Results} \textbf{C}). 

Specifically, the F1 values of sensor-dependent NNs for thirteen participants were as follows: FSR+PEF = 51.00\%, orientation = 58.00\%, acceleration = 63.00\%, and microphones = 44.00\%. 
For subjects 2, 3, 6, 9 and 11, the ensemble model increased performance by up to 28.00\% of the F1 result.
In some cases, the ensemble model results were limited to the best sensor type NN (inertial sensing) performance in participants (7 and 8), indicating that the ensemble model did not improve the performance in these particular cases. 

In \cref{fig:Results} \textbf{C}, the most evident misclassification is between sadness and anger with up to 17.00\% confusion.  
This could be the consequence of the misinterpreted eyebrow movement while doing sadness; instead of eyebrows up, many volunteers moved their eyebrows down and the intensity of the gesture.
A 13.00\% confusion happens with the faces of anger and disgust, which are depicted as similar in \cref{fig:FacialSignals}.  
Another relevant misclassification (10.00\%) occurs with surprise and joy gestures, which are faces distinguished by the mouth movement.
The recognition value of 98.00\% of the neutral face (null class) indicates the high specificity of the design, making null class recognition a suitable candidate for automatic data segmentation in real time. 

The AMMG sensing modality has the weakest performance. 
Multiple reasons can lead to this result; the facial expression usually involves audio patterns in a natural setting, which might not be present in a mimicked experiment. 
The audio analysis in this work only considers the MFCCS with 13 filters to speed up the NN converging time but overlooks the spectral information in the entire spectrum. 
Due to the small dataset, more elaborated image processing techniques were not explored.
In the future, it is highly recommended to do data augmentation and transfer learning to exploit the audio information completely. 

\begin{figure*}[!t]
    \includegraphics[width=\textwidth]{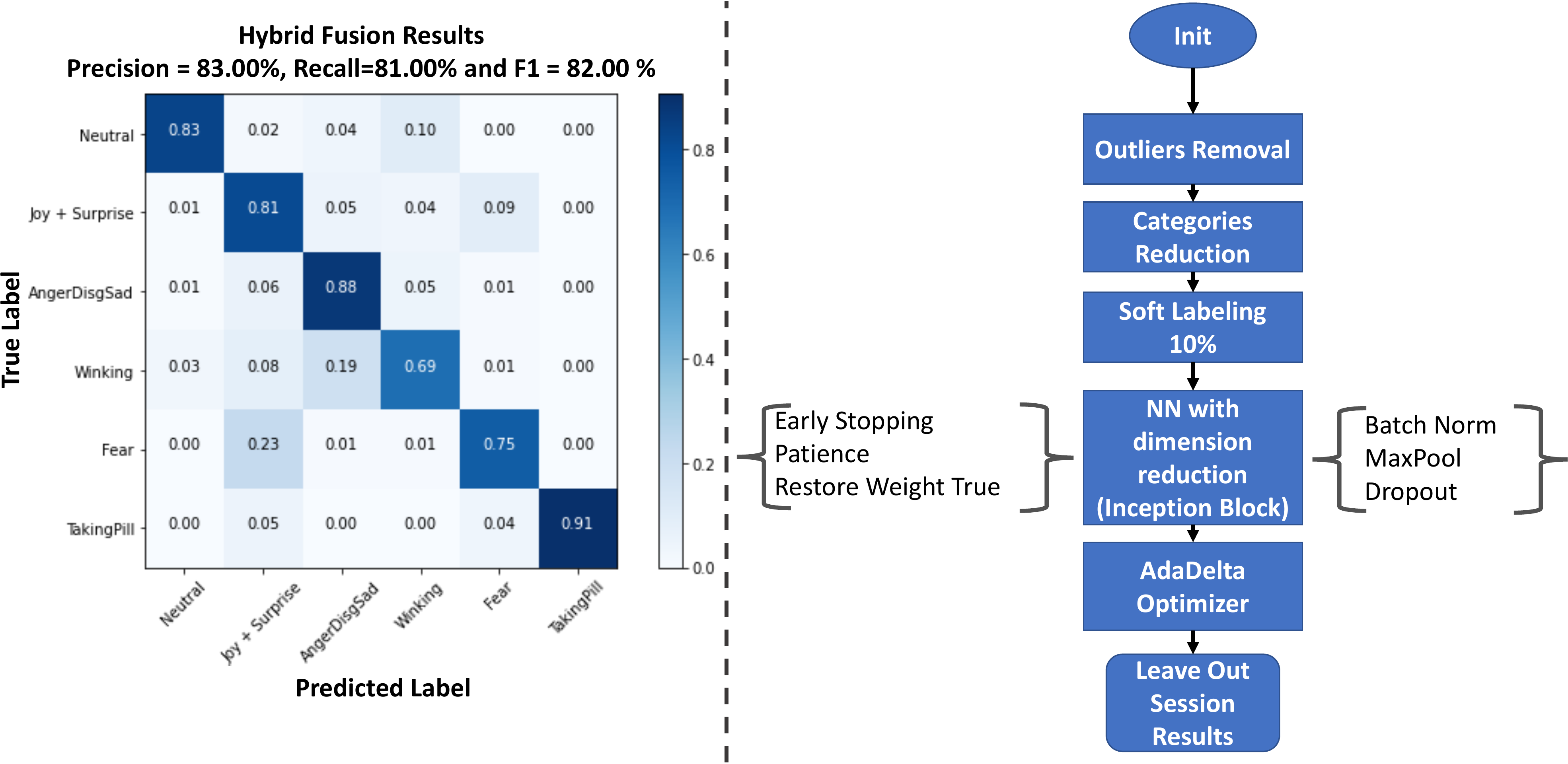}
    \caption{\textbf{Left} Recall Results of Hybrid Fusion Model Leave Out Session for the Eight Best Imitators F1 = 82\%. \textbf{Right} Data Analysis Techniques Applied in the Hybrid Fusion Modeling Pipeline.}
    \label{fig:ResultsHybrid}
 \end{figure*}
 \cref{fig:ResultsHybrid} \textbf{Left} shows the performance of the hybrid fusion model using a leave-out session evaluation scheme. 
The result has an F1-score = 82 \%, which is the average of five iterations/sessions.
The main confusions are between the pairs; "Fear-Joy + Surprise" and "Winking-Anger + Disgust + Sadness", with 23 \% and 19 \%, respectively.
It is relevant to notice that the confused classes have similar eyebrow movements. 
The first pair consists of expressions dominated by up-eyebrow movements, and down-eyebrow movements mainly dominate the second. 
However, the confusion is in the categories with fewer instances. 
Due to the merged categories, the model can learn the distinction in the complementary classes with recall above 80.00 \% for the case of "Joy + Surprise" and up to 88.00 \% recall for the case of "Anger + Disgust + Sadness.
Despite our small dataset, an F1 score over 80 \% is an encouraging outcome. 
The outcome can be improved if more data is fed to the NN. 

The result of the hybrid fusion is the output of the data analysis pipeline depicted in \cref{fig:ResultsHybrid} \textbf{Right}. 
Thus, a direct comparison between late and hybrid fusion is impossible. 
Still, the low complexity and parallelism of the modular late fusion, one sensor-dependent NN per MCU, make it our preferable option to be used in embedded devices. 
Additionally, the hybrid fusion employs 2.35x more parameters than the early fusion to exploit the correlations between the sensing modalities.
In the future, the hybrid model results could be improved if more data were supplied to the NN.
It is worth exploring NN partitioning techniques to solve the lack of parallelism in the hybrid fusion model.
In \cite{DistributedCNN}, the authors proposed to leverage the power of multiple embedded/edge devices to run a large CNN. 
They proposed a framework that automatically partitions a CNN model into sub-models and generates the code for the execution of these sub-models on multiple edge devices (possibly heterogeneous) while supporting the exploitation of parallelism between and within edge devices.

\subsection{Discussion}
It is important to note that these results are based on a facial expression imitation experiment with participants from different cultures with no acting experience.
Although each participant interpreted facial activity in her/his own way, we found that complementary sensing modalities could detect patterns.

Due to the design of our prototype, our solution has several limitations. 
One of them is that when using a sports cap as a wearable accessory, cheek and mouth movements are captured indirectly and without a complete map of the facial activity. 
The size of the custom-built electronics board and the chosen battery make the system heavy for real-life daily use.
Downsizing the prototype board is inevitable to test the idea in realistic scenarios with natural facial expressions.
The distributed sensing modalities in the frontalis and temporalis muscles may also lead to variations in MMG signal strength. 
A definitive statement about the best sensor modality could only be made if the placement of the sensors is the same, but this would be at the expense of user comfort, at least with our current hardware.
A possible alternative is to stack (sandwich) all the sensors in a miniature circuit prototype; in this way, a quantitative comparison can be made between the sensing modality and the type of activity. 

On the other hand, based on our encouraging offline results, real-time implementation of the models is a reasonable next step. 
However, at the current stage, the system relies on segmenting the entire facial gesture (start/end point) to do the recognition. 
In the future, a plan to test the idea of online inference is to use the FSR gradient as a trigger for automatic data segmentation (unrest state recognition) and then proceed to inference. 

In general, the sensing approaches and the modular technique can be applied to many different fields, such as psychology, to monitor students' learning process in classrooms \cite{FacialCamKidClassroom}. 
Nevertheless, a clinician/cognitive expert will still need to redesign the experiment to adapt it to the psychological requirements.  
Overall, our design goal is to explore the system's feasibility in detecting facial muscle movements.
\section{Conclusion and Outlook}
\label{sec:conclusion}
This work presents a privacy-preserving, low-memory, low-power consumption, and unobtrusive alternative system for facial expression detection.
Although the design uses microphones, it only uses them to capture the sound of muscles while facial movements are performed, so no voice or ambient sound will affect the privacy requirement. 
Our system has demonstrated the ability to detect facial expressions using non-camera sensors mounted on a sports cap.
In addition, the system works on participants of both sexes and from different cultures, demonstrating the inclusivity and generalizability of the approach.
In the individual and cross-user results, the remounting of the sports cap was accountable; hence, our results are robust against everyday re-wearing in wearables. 
The results indicate that a multimodal approach based on the proposed sensors is well-suited for recognizing facial activities with an unobtrusive wearable sensor system. 

Compared to the solutions based on our single sensing models, our ensemble approach provides a performance improvement of 16 \%. 
IMU and PMMG were the two sensing methods that contributed the most significantly to our results. 
An interesting question for future studies is how to extract more information from AMMG while maintaining a low-memory, privacy-friendly design. 
In particular, data augmentation and transfer learning methods could be used to understand the AMMG better. 
Also, it would be necessary to consider different sensors' placement around the head for a detailed comparison between sensing methods. 
One promising approach will be to design a miniature prototype as a stack of sensor units consisting of pressure sensors, piezoelectric sensors, a mechanical microphone, and an inertial measurement unit.
The sensor stack would simplify the connection between the sensor to be selected and the types of muscle activities to be monitored. 

Our results are from an offline analysis. 
Therefore, obstacles may appear to the system's deployment, such as environmental sounds, power consumption, mobile scenarios, and user comfort.
In the future, a miniaturized version of the system is desirable to perform experiments with more people and in realistic (out-of-office) scenarios.
The use of an entire gesture-instance data processing technique has the advantage of lighter and faster models (due to its simplicity). 
At the same time, it makes the design dependent on automatic segmentation techniques to determine the start/end point of a facial gesture for the case of real-time gesture recognition. 
For automatic gesture segmentation, it is possible to select one or a combination of the following paths; the FSR pressure data gradient can be used as a trigger to detect a person's unrest state and then proceed to signal the start/stop of data collection to make the prediction. 
The FSR gradient option is also suitable for reducing power consumption.  
The solution can be combined with a weighted belief system and use the PEF gradient with the FSR; it is necessary to consider that the PEF is susceptible to the motion of the worn accessory, a sports cap, in our case. 
The FSR-PEF disturbance detection system can be the first step in a hierarchical procedure in which the FSR-PEF model is used to detect the null class (83.00 \% recall for thirteen volunteers). 
Then the ensemble/fusion model is activated.

We employ seven validated facial expressions (Warsaw Photoset) in our work to compare our results with future solutions. 
Although a fair comparison with related work is negatively affected by many research papers using personalized facial/head activities, we believe our results are competitive with the state of the art. 
Several challenges remain before we can deploy the system immediately in embedded devices, but with the results of this work, we believe in the idea's potential. 

	\section*{Acknowledgments}
    The research reported in this paper was supported by the BMBF (German Federal Ministry of Education and Research) in the project SocialWear. 
	
	\bibliography{mybibfile}

\end{document}